# Statistical Feature Combination for the Evaluation of Game Positions

**Michael Buro**  MIC@RESEARCH.NJ.NEC.COM
*NEC Research Institute*
*4 Independence Way*
*Princeton NJ 08540 U.S.A.*

## Abstract

This article describes an application of three well–known statistical methods in the field of game–tree search: using a large number of classified Othello positions, feature weights for evaluation functions with a game–phase–independent meaning are estimated by means of logistic regression, Fisher's linear discriminant, and the quadratic discriminant function for normally distributed features. Thereafter, the playing strengths are compared by means of tournaments between the resulting versions of a world–class Othello program. In this application, logistic regression — which is used here for the first time in the context of game playing — leads to better results than the other approaches.

## 1. Introduction

Programs playing games like chess, draughts, or Othello use evaluation functions to estimate the players' winning chances in positions at the leaves of game–trees. These values are propagated to the root according to the NegaMax principle in order to choose a move in the root position which leads to the highest score. Normally, evaluation functions combine features that measure properties of the position correlated with the winning chance, such as material in chess or mobility in Othello. Most popular are quickly computable linear feature combinations. In the early days of game programming, the feature weights were chosen intuitively and improved in a manual hill–climbing process until the programmer's patience gave out. This technique is laborious. Samuel (1959,1967) was the first to describe a method for automatic improvement of evaluation function parameters. Since then many approaches have been investigated. Two main strategies can be distinguished:

**Move adaptation:** Evaluation function parameters are tuned to maximize the frequency with which searches yield moves that occur in the lists of moves belonging to training positions. The idea is to get the program to mimic experts' moves.

**Value adaptation:** Given a set of labelled example positions, parameters are determined such that the evaluation function fits a specific model. For instance, evaluation functions can be constructed in this way to predict the final game result.

In move adaptation, proposed for instance by Marsland (1985), v.d. Meulen (1989), and Mysliwietz (1994), a linear feature combination has two degrees of freedom: it can be multiplied by a positive constant and any constant can be added to it without changing the move decision. If the evaluation function depends on the game phase, and positions from different phases are compared (for example within the framework of selective extensions or opening book play), these constants must be chosen suitably. Because evaluation functions optimized by move adaptation for the moment have no global interpretation, a solution of this problem is not obvious. Schaeffer et al. (1992) presented an *ad hoc* and game–specific approach.

In this respect, value adaptation is more promising. Here, evaluations from different phases are comparable if the example position labels have a phase–independent meaning. Mitchell (1984) labelled Othello positions occurring in a game with the final game result in the form of the disc differential and tried to approximate these values using a linear combination of features. Since a regression was used to determine the weights, it was also possible to investigate the features' statistical relevance. Another statistical approach for value adaptation was used by Lee & Mahajan (1988): example positions were classified as a win or loss for the side to move and — assuming the features to be multivariate normal — a quadratic discriminant function was used to predict the winning probability. This technique ensures the desired comparability and applies also to games without win degrees, i.e. that only know wins, draws, and losses.

Besides these classical approaches which heavily rely on given feature sets, in recent years artificial neural networks (ANNs) have been trained for evaluating game positions. For instance, Moriarty & Miikkulainen (1993) used genetic algorithms to evolve both the topology and weights of ANNs in order to learn Othello concepts by means of tournaments against fixed programs. After discovering the concept of mobility, their best 1–ply ANN–player was able to win 70% of the games against a 3–ply brute–force program that used an evaluation function without mobility features. The most important contribution in this field is by Tesauro (1992,1994,1995). Using temporal difference learning (Sutton, 1988) for updating the weights, his ANNs learned to evaluate backgammon positions at master level by means of self–play. Tesauro conjectured the stochastic nature of backgammon to be responsible for the success of this approach. Though several researchers obtained encouraging preliminary results applying Tesauro's learning procedure to deterministic games, this work has not yet led to strong tournament programs for tactical games such as Awari, draughts, Othello, or chess, that allow deep searches and for which powerful and quickly computable evaluation functions are known. It might be that due to tactics for these games more knowledgeable but slower evaluation functions are not necessarily more accurate than relatively simple and faster evaluation functions in conjunction with deeper searches.

In what follows, three well–known statistical models — namely the quadratic discrimination function for normally distributed features, Fisher's linear discriminant, and logistic regression — are described for the evaluation of game positions in the context of value adaptation. Thereafter, it is shown how example positions for parameter estimation were generated. Finally, the playing strengths of three versions of a world–class Othello program[1] — LOGISTELLO — equipped with the resulting evaluation functions are compared in order to determine the strongest tournament player. It turns out that quadratic feature combinations do not necessarily lead to stronger programs than linear combinations, and that logistic regression gives the best results in this application.

## 2. Statistical Feature Combination

The formal basis of statistical feature combination for position evaluation can be stated as follows:

---

1. Since its appearance in October 1993 it won twelve of the 14 international tournaments it played.

STATISTICAL FEATURE COMBINATION

- $\Omega$ is the set of positions to evaluate.

- $Y : \Omega \to \{\mathcal{L}, \mathcal{W}\}$ classifies positions as a loss or win for the player to move, assuming optimal play by both sides. Draws can be handled in the manner outlined in Section 4.

- $X_1, \ldots, X_n : \Omega \to \mathbb{R}$ are the features.

- The evaluation of a position $\omega \in \Omega$ with $\boldsymbol{x} = (X_1, \ldots, X_n)(\omega)$ is the conditional winning probability

$$V(\omega) = P(Y = \mathcal{W} \mid (X_1, \ldots, X_n) = \boldsymbol{x}) =: P(\mathcal{W} \mid \boldsymbol{x}).$$

- There are $N$ classified example positions $\omega_1, \ldots, \omega_N \in \Omega$ available with $\boldsymbol{x}_i = (X_1, \ldots, X_n)(\omega_i)$ and $y_i = Y(\omega_i)$.

In the following subsections models which express $P(\mathcal{W} \mid \boldsymbol{x})$ as a function of linear or quadratic feature combinations are briefly introduced in a way that is sufficient for practical purposes. Good introductions and further theoretical details are given for instance by Duda & Hart (1973), Hand (1981), Agresti (1990), and McCullagh & Nelder (1989). Both Fisher's classical method and logistic regression are used here to model $P(\mathcal{W} \mid \boldsymbol{x})$ for the first time; the quadratic discriminant function has been used by Lee & Mahajan (1988), however, without considering Fisher's discriminant first.

## 2.1 Discriminant Functions for Normally Distributed Features

Bayes' rule gives

$$\begin{aligned} P(\mathcal{W} \mid \boldsymbol{x}) &= \frac{p(\boldsymbol{x} \mid \mathcal{W}) P(\mathcal{W})}{p(\boldsymbol{x})} = \frac{p(\boldsymbol{x} \mid \mathcal{W}) P(\mathcal{W})}{p(\boldsymbol{x} \mid \mathcal{W}) P(\mathcal{W}) + p(\boldsymbol{x} \mid \mathcal{L}) P(\mathcal{L})} \\ &= \left(1 + \frac{p(\boldsymbol{x} \mid \mathcal{L}) P(\mathcal{L})}{p(\boldsymbol{x} \mid \mathcal{W}) P(\mathcal{W})}\right)^{-1}, \end{aligned}$$

where $p(\boldsymbol{x} \mid C)$ is the features' conditional density function and $P(C)$ is the a priori probability of class $C \in \{\mathcal{L}, \mathcal{W}\}$. In the case that the a priori probablities are equal, and the features are multivariate normally distributed within each class, i.e.

$$p(\boldsymbol{x} \mid C) = (2\pi)^{-n/2} |\Sigma_C|^{-1/2} \exp\left\{-\tfrac{1}{2}(\boldsymbol{x} - \boldsymbol{\mu}_C) \Sigma_C^{-1} (\boldsymbol{x} - \boldsymbol{\mu}_C)'\right\}$$

with mean vector $\boldsymbol{\mu}_C$ and covariance matrix $\Sigma_C$ for $C \in \{\mathcal{W}, \mathcal{L}\}$, it follows

$$P(\mathcal{W} \mid \boldsymbol{x}) = \frac{1}{1 + \exp(-f(\boldsymbol{x}))},$$

where $f$ is the following quadratic discriminant function:

$$\begin{aligned} f(\boldsymbol{x}) &= -\Big\{\tfrac{1}{2}\boldsymbol{x}\left(\Sigma_\mathcal{W}^{-1} - \Sigma_\mathcal{L}^{-1}\right)\boldsymbol{x}' + \left(\boldsymbol{\mu}_\mathcal{L} \Sigma_\mathcal{L}^{-1} - \boldsymbol{\mu}_\mathcal{W} \Sigma_\mathcal{W}^{-1}\right)\boldsymbol{x}' + \\ &\quad \tfrac{1}{2}\left(\boldsymbol{\mu}_\mathcal{W} \Sigma_\mathcal{W}^{-1} \boldsymbol{\mu}_\mathcal{W}' - \boldsymbol{\mu}_\mathcal{L} \Sigma_\mathcal{L}^{-1} \boldsymbol{\mu}_\mathcal{L}' + \log|\Sigma_\mathcal{W}| - \log|\Sigma_\mathcal{L}|\right)\Big\}. \end{aligned}$$

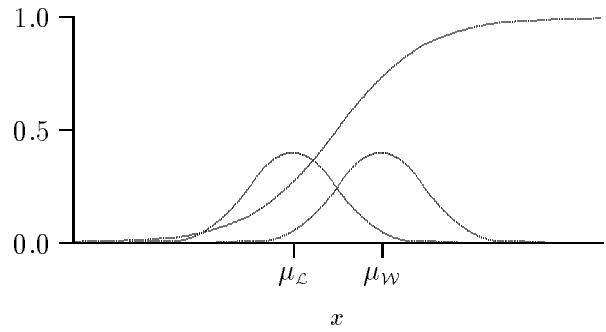

Figure 1: Conditional densities and winning probability

If the covariance matrices are equal ($=\Sigma$), the expression can be simplified to a linear function:

$$f(\boldsymbol{x}) \;\;=\;\; (\boldsymbol{\mu}_{\mathcal{W}} - \boldsymbol{\mu}_{\mathcal{L}})\Sigma^{-1}\{\boldsymbol{x} - (\boldsymbol{\mu}_{\mathcal{L}} + \boldsymbol{\mu}_{\mathcal{W}})/2\}'.$$

Interestingly, this function is also a solution to the problem of finding a linear transformation which maximizes the ratio of the squared sample mean distance to the sum of the within–class sample variances after transformation. Therefore, it has good separator properties even if the features are not normally distributed. This is called Fisher's linear discriminant. Figure 1 illustrates the relation between the conditional densities and the winning probability.

The maximum likelihood (ML) parameter estimates are

$$\hat{\boldsymbol{\mu}}_C \;\;=\;\; \frac{1}{|I_C|}\sum_{i \in I_C} \boldsymbol{x}_i$$

$$\hat{\Sigma}_C \;\;=\;\; \frac{1}{|I_C|}\sum_{i \in I_C} (\boldsymbol{x}_i - \hat{\boldsymbol{\mu}}_C)'(\boldsymbol{x}_i - \hat{\boldsymbol{\mu}}_C)$$

with $I_C = \{i \mid y_i = C\}$. If the covariance matrices are equal,

$$\hat{\Sigma} = \frac{1}{|I_{\mathcal{W}}| + |I_{\mathcal{L}}|} \sum_{C \in \{\mathcal{L},\mathcal{W}\}} \sum_{i \in I_C} (\boldsymbol{x}_i - \hat{\boldsymbol{\mu}}_C)'(\boldsymbol{x}_i - \hat{\boldsymbol{\mu}}_C).$$

### 2.2 Logistic Regression

In logistic regression the conditional winning probability $P(\mathcal{W} \mid \boldsymbol{x})$ depends on a linear combination of the $x_i$. Here, $X_1 \equiv 1$ is assumed in order to be able to model constant offsets. The simple approach $P(\mathcal{W} \mid \boldsymbol{x}) = \boldsymbol{x}\boldsymbol{\beta}$ using a parameter column vector $\boldsymbol{\beta}$ is unusable because $\boldsymbol{x}\boldsymbol{\beta} \in [0,1]$ cannot be guaranteed generally. This requirement can be fulfilled by means of a link–function $g : (0,1) \to \mathbb{R}$ according to $g(P(\mathcal{W} \mid \boldsymbol{x})) = \boldsymbol{x}\boldsymbol{\beta}$. Figure 2 shows a typical nonlinear relation between the winning probability and one feature. Since the probability is usually a monotone increasing function of the features, $g$ should satisfy $\lim_{x \to 0+} g(x) = -\infty$ and $\lim_{x \to 1-} g(x) = +\infty$. The link–function $g(t) = \text{logit}(t) := \log(t/(1-t))$ has these properties. Using $g = \text{logit}$, since $g^{-1}(x) = \{1 + \exp(-x)\}^{-1}$, it follows that

$$P(\mathcal{W} \mid \boldsymbol{x}) = \frac{1}{1 + \exp(-\boldsymbol{x}\boldsymbol{\beta})}.$$

Hence, the winning probability has the same shape as for discriminant analysis. But logistic regression does not require the features to be multivariate normal; even the use of very discrete features is possible.

Again the parameter vector $\boldsymbol{\beta}$ can be estimated using the ML approach. Unfortunately, in this case it is necessary to solve a system of nonlinear equations. In what follows, a known solving approach will be briefly described (cf. Agresti, 1990; McCullagh & Nelder, 1989).

In order to ensure convergence of the iterative algorithm given below, it is necessary to slightly generalize our model: from now on $y_i$ is the observed value of random variable $Y_i = \sum_{j=1}^{n_i} Y_{i,j}$, where the $Y_{i,j} : \Omega \to \{0,1\}$ have mean $\pi_i = \{1 + \exp(-\boldsymbol{x}_i\boldsymbol{\beta})\}^{-1}$ and are stochastically independent. This definition includes the old model ($n_i = 1$ and $y_i \in \{0,1\}$). The likelihood function $L(\boldsymbol{\beta})$, which is a probability density, measures how likely it is to see the realization $\boldsymbol{y}$ of the stochastically independent random variables $Y_i$, if $\boldsymbol{\beta}$ is the true parameter vector. In order to maximize $L$, it suffices to consider $\log(L)$:

$$\begin{aligned}\log(L(\boldsymbol{\beta})) &= \log\Big(\prod_{i=1}^{N} \pi_i^{y_i}(1-\pi_i)^{n_i-y_i}\Big) = \sum_{i=1}^{N} y_i \log \pi_i + (n_i - y_i)\log(1-\pi_i) \\ &= \sum_{j=1}^{n}\Big(\sum_{i=1}^{N} y_i x_{ij}\Big)\beta_j - \sum_{i=1}^{N} n_i \log\Big[1 + \exp\Big(\sum_{j=1}^{n} x_{ij}\beta_j\Big)\Big].\end{aligned}$$

This function is twice differentiable, is strictly concave up to rare border cases, and has a unique maximum location if $0 < y_i < n_i$ for all $i$ (cf. Wedderburn, 1976) that can be iteratively found using the Newton–Raphson method as follows:

$$\hat{\boldsymbol{\beta}}^{(t+1)} = (\boldsymbol{X}'\boldsymbol{\Delta}^{(t)}\boldsymbol{X})^{-1}\boldsymbol{X}'\boldsymbol{\Delta}^{(t)}\boldsymbol{z}^{(t)}$$

with

- the $(N \times n)$–matrix $\boldsymbol{X}$ built from the $\boldsymbol{x}_i$,

- $\boldsymbol{\Delta}^{(t)} = \mathbf{diag}[n_i\hat{\pi}_i^{(t)}(1-\hat{\pi}_i^{(t)})]$, $\hat{\pi}_i^{(t)} = \Big\{1 + \exp\Big(-\sum_{j=1}^{n} x_{ij}\hat{\beta}_j^{(t)}\Big)\Big\}^{-1}$, and

- $z_i^{(t)} = \log\Big(\dfrac{\hat{\pi}_i^{(t)}}{1-\hat{\pi}_i^{(t)}}\Big) + \dfrac{y_i - n_i\hat{\pi}_i^{(t)}}{n_i\hat{\pi}_i^{(t)}(1-\hat{\pi}_i^{(t)})}$.

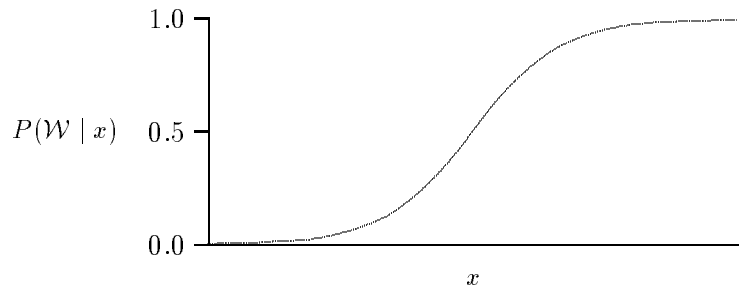

Figure 2: Typical shape of the winning probability

Starting with $\hat{\pi}_i^{(0)} = (y_i+1/2)/(n_i+1)$, the ML estimate $\hat{\boldsymbol{\beta}}$ may usually be computed with high accuracy within a few steps since the method is quadratically convergent and relatively robust with respect to the choice of the starting vector. Unfortunately, if there is an $i$ with $y_i = 0$ or $y_i = n_i$ the estimates might not converge. But our original model can be approximated, for instance, by setting $n_i = 100$ and $y_i = 1$ or $99$, depending on whether the position in question is lost or won.

## 3. Generation and Classification of Example Positions

Value adaptation requires labelled example positions. Here, some problems arise. First of all, for most nontrivial games only endgame positions can be classified correctly as won, drawn, or lost; for opening and midgame positions optimal play is out of reach due to the lack of game knowledge and time constraints. Furthermore, the example positions should contain significant feature variance since otherwise no discrimination is possible. Hence, it is problematic to use only high level games — which might be the first idea — since good players and programs know the relevant features and try to maximize them during a game. Therefore, these features tend to be constant most of the time and statistical methods would assign only small weights to them. As a final difficulty, estimating parameters accurately for different game phases requires many positions.

A pragmatic "solution" to these problems is indicated in Figure 3: over a period of two years, about 60,000 Othello games[2] were played by early versions of LOGISTELLO and Igor Đurđanović's program REV.[3] Feature variance was ensured by examining all openings of length seven which led mostly to unbalanced starting positions. Since early program versions were used which had only 5–10 minutes thinking time, the games, though well played most of the time, are not error free. In some cases even big mistakes occurred in which, for example, one side fell into a corner losing trap[4] caused by a lack of look–ahead. But without these errors, no reasonable weight estimation of principal features (such as corner possession in Othello) is possible as explained above. Following Lee & Mahajan (1988), all positions were then classified by the final game results. This approach is problematic because the classification reliability decreases from the endgame to the opening phase due to player mistakes. To reduce this effect, early outcome searches were performed for solving Othello positions 20 moves before game end. Furthermore, from time to time the game database was searched for "obvious" errors using new program versions and longer searches to correct these games. Since in this process many lines of play were repeated, the misclassification rate was further reduced by propagating the game results from the leaves to the root of the game–tree, which had been built from all games according to the NegaMax principle. In this way the classification of a position depends on that of *all* examined successors and is therefore more reliable.

The proposed classification method is relatively fast and allows us to label many positions in a reasonable time (on average about 42 new positions in 10–20 minutes). In addition to

---

2. The game file can be obtained via anonymous ftp.
   (ftp.uni-paderborn.de/unix/othello/misc/database.zip)
3. A brief description of both programs is given in the help pages of the Internet Othello Server.
   (telnet faust.uni-paderborn.de 5000)
4. In its implications this can be compared with losing material for nothing in chess.

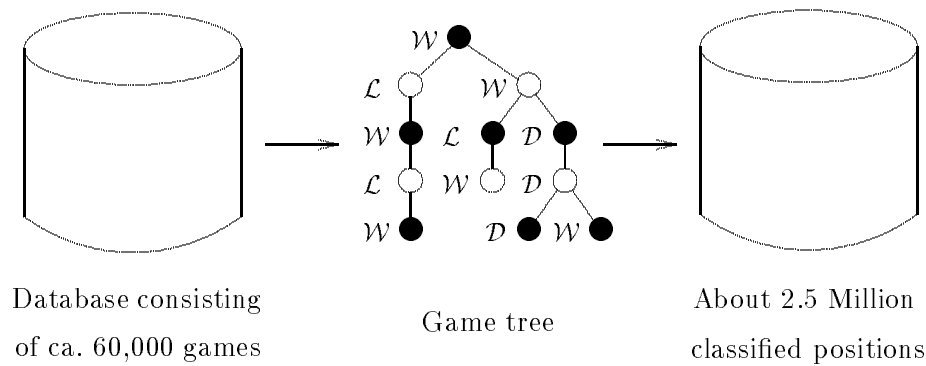

Database consisting of ca. 60,000 games    Game tree    About 2.5 Million classified positions ensuring an accurate parameter estimation even for different game phases (which is indicated by small parameter confidence intervals), this method enabled us to develop new pattern features for Othello based on estimating the winning probability conditioned upon occurrence of sub–configurations, like edge or diagonal instances, of the board.[5]

Figure 3: The classification process

## 4. Parameter Estimation and Playing Strength Comparison

Although only about 5% of the example positions were labelled as drawn, it was decided to use them for parameter estimation since these positions give exact information about feature balancing. A natural way to handle drawn positions within the statistical evaluation framework considered here is to define the winning probability to be 1/2 in this case. For this extension the logistic regression parameters can easily determined by setting $y_i = n_i/2$ in case of a draw. Alternatively, doubling won or lost positions and incorporating drawn positions once as won and once as lost leads to the same estimate because both log likelihood functions are equal up to a constant factor. The latter technique was used for fitting the other models.

Previous experiments showed that the parameters depend on the game phase, for which disc count is an adequate measure in Othello. So the example positions were grouped according to the number of discs on the board, and adjacent groups were used for parameter estimation in order to smooth the data and to ensure almost equal numbers of won and lost positions.

The success of the Othello program **BILL** described by Lee & Mahajan (1990) shows that in Othello table–based features can be quite effective. For instance, the important edge structure can be quickly evaluated by adding four pre–computed edge evaluations which are stored in a table. All 13 features used by **LOGISTELLO** are table–based. They fall into two groups: in the first group pattern instances including the horizontal, vertical, and most diagonal lines of the board are evaluated while in the second group two mobility measures are computed.[5]

After parameter estimation for the three described models, tournaments between the players **QUAD** (which uses the quadratic discriminant function for normally distributed features), FISHER, and LOG were played in order to determine the best tournament player. Starting with 100 nearly even opening positions with 14 discs (i.e. before move 11) from LOGISTELLO's opening book, each game and its return game with colours reversed was played.[6] In the opening and midgame phase all program versions performed their usual iterative deepening NegaScout searches (Reinefeld, 1983) with a selective corner quiescence search extension. Endgame positions with about 22 empty squares were solved by win–draw–loss searches. There was no pattern learning during the tournaments, and the facility to think on opponent's time was turned off in order to speed up the tournaments which were run in parallel on seven SUN SPARC–10 workstations.

---

5. Details are given by Buro (1994). The postscript file of this thesis can be obtained via anonymous ftp.

| Pairing | Time per game (Minutes) | Result (Win−Draw−Loss) | Winning Percentage |
|---|---|---|---|
| LOG − QUAD | 30 − 30 | 116 − 15 − 69 | 61.8% |
| FISHER − QUAD | 30 − 30 | 112 − 15 − 73 | 59.8% |
| LOG − FISHER | 30 − 30 | 93 − 35 − 72 | 55.3% |
| LOG − FISHER | 30 − 36 | 86 − 24 − 90 | 49.0% |
| LOG − QUAD | 30 − 38 | 93 − 33 − 74 | 54.8% |
| FISHER − QUAD | 30 − 38 | 84 − 30 − 86 | 49.5% |
| LOG − QUAD | 30 − 45 | 88 − 26 − 86 | 50.5% |

Table 1: Tournament results

Applying a conservative statistical test[5] it can be seen that all results listed in Table 1 stating a winning percentage greater than 59% are statistically significant at the 5% level. The first two results show a clear advantage for the linear combinations under normal tournament conditions (30 minutes per player per game). Furthermore, since LOG outperforms FISHER the features would not seem to be even approximately normally distributed. Here lies the advantage of logistic regression: even very discrete features like castling status in chess or parity in Othello can be used.

Further tournaments were played with more time for the weaker players FISHER and QUAD in order to determine the time factors which lead to an equal playing strength. As shown in Table 1 FISHER reaches LOG's strength if it is given about 20% more time, and QUAD needs about 50% more time to compete with LOG. With LOGISTELLO's optimized implementation, the search speed when using the quadratic combination is still about 20% slower than that with the linear combination. Thus, giving QUAD 25% more time ($1/(1 - 0.2) = 1.25$) balances the total number of nodes searched during a game. But even with this timing, LOG is stronger than QUAD, and FISHER can still compete with it. All in all, the quadratic combination is not only slower than the linear combination, but it also has no better discrimination properties. Indeed, a look at the estimated covariance matrices of each class revealed that they are almost equal, and therefore a better evaluation quality than that of Fisher's linear discriminant could not be expected.

---

6. LOG's 11–ply evaluation of these positions lies in the range $[-0.4, +0.4]$ which corresponds to winning probabilities in the range $[0.4, 0.6]$. Only nearly even starting positions should be used to compare programs of similar playing strength since in clear positions the colour determines the winner and the winning percentage would be 50% even if one player is stronger. Of the 100 starting positions only six always led to game pairs with a balanced score.

## 5. Discussion

In this paper three statistical approaches for modelling evaluation functions with a game–phase–independent meaning have been presented and compared empirically using a world–class Othello program. Quadratic feature combinations do not necessarily lead to stronger programs than linear combinations since the evaluation speed can drop significantly. Of course, this effect depends on the number of features used and their evaluation speed: if only a few features are used or if it takes a long time to evaluate them, then the playing strength differences cannot be explained by different speeds because in this case the evaluation times are almost equal. In any case, before using quadratic combinations the covariance matrices should be compared; if they are (almost) equal, the quadratic terms can be omitted and Fisher's linear discriminant can be used. Therefore, the motivations of Lee & Mahajan (1988) need refinement, since an existing feature correlation does not necessarily justify the use of nonlinear combinations. Generally, possibly more accurate nonlinear feature combinations (such as ANNs) should be compared to simpler but faster approaches *in practice*, since their use does not always guarantee a greater playing strength.

Besides linear regression and discriminant analysis, logistic regression has proven to be a suitable tool for the construction of evaluation functions with a global interpretation. The drawback, that for parameter estimation a system of nonlinear equations has to be solved, is more than compensated for by the higher quality of the evaluation function in comparison to the other approaches, since in this application the parameters have to be determined only once. The current tournament version of **LOGISTELLO** uses feature weights estimated by means of logistic regression and profits from the comparability of evaluations from different game phases which is ensured by the use of value adaptation. As a result it is possible to perform selective searches in which values from different game phases are compared; moreover, values from the opening can be compared even with late midgame values in order to find promising move alternatives in the program's opening book (Buro 1994,1995). In this sense, value comparability is a cornerstone of **LOGISTELLO**'s strength.

### Acknowledgements

I wish to thank my wife Karen for competently answering many of my statistical questions. I also thank my colleague Igor Đurđanović for many fruitful discussions which have led to considerable improvements of our Othello programs. Furthermore, I am grateful to Colin Springer, Richard E. Korf, and the anonymous referees for their useful suggestions on earlier versions of this paper, which helped improve both the presentation and the contents.